\documentclass[11pt,a4paper]{article}
\usepackage{microtype}
\usepackage{graphicx}
\usepackage{booktabs}
\usepackage[english]{babel}
\usepackage[utf8x]{inputenc}
\usepackage[T1]{fontenc}
\usepackage{amsfonts}
\usepackage{subcaption}
\usepackage[font={large}]{caption}
\usepackage[numbers]{natbib}
\usepackage{amsmath}
\usepackage{mathabx}
\usepackage{graphicx}
\usepackage{adjustbox}
\usepackage[colorinlistoftodos]{todonotes}
\usepackage[colorlinks=true, allcolors=blue]{hyperref}
\usepackage{amsthm}
\usepackage{lipsum}

\theoremstyle{definition}

\usepackage{hyperref}

\usepackage{authblk}

\usepackage[left=3cm, right=3cm, top=3cm]{geometry} 

\usepackage{tikz}
\usetikzlibrary{arrows}
\usepackage{wrapfig}
\usepackage{multirow}
\usepackage[algo2e,ruled]{algorithm2e}
\usepackage{algorithm}

\tikzset{
  treenode/.style = {align=center, inner sep=0pt, text centered, font=\sffamily},
  arn_n/.style = {treenode, circle, black, font=\sffamily\bfseries, draw=black, text width=2em}
}

\title{Active Refinement for Multi-Label Learning:\\ A Pseudo-Label Approach}
\author[1]{Cheng-Yu Hsieh\thanks{r05922048@ntu.edu.tw}}
\author[1]{Wei-I Lin\thanks{r10922076@ntu.edu.tw}}
\author[2,3]{Miao Xu\thanks{miao.xu@uq.edu.au}}
\author[3]{Gang Niu\thanks{gang.niu.ml@gmail.com}}
\author[1]{Hsuan-Tien Lin\thanks{htlin@csie.ntu.edu.tw}}
\author[3,4]{Masashi Sugiyama\thanks{sugi@k.u-tokyo.ac.jp}}

\affil[1]{National Taiwan University}
\affil[2]{The University of Queensland}
\affil[3]{RIKEN Center for Advanced Intelligence Project}
\affil[4]{The University of Tokyo}

\date{}
\begin{document}
\maketitle
  
    \newcommand{\func}{\mathbf{ f}}
    \newcommand{\ex}{\Phi}
    \newcommand{\exm}{\widebar{\Phi}}
    \newcommand{\x}{\mathbf{x}}
    \newcommand{\y}{\mathbf{ y}}
    \newcommand{\z}{\mathbf{z}}
    \newcommand{\E}{\mathbb{E}}
    \newcommand{\ep}{ r}
    \newcommand{\s}{S}
    \newcommand{\T}{\mathbf{ t}}
    \newcommand{\A}{\mathbf{a}}
    \newcommand{\uu}{\mathbf{u}}
    \def\Smax{S_{\textsc{MAX}}}
    \def\Savg{S_{\textsc{AVG}}}
    \def\Pmax{P_{\textsc{MAX}}}
    \def\Pavg{P_{\textsc{AVG}}}
    \def\Pxxx{P}    
    \def\Sxxx{S}
    \newcommand\Set[2]{\{\,#1\mid#2\,\}}

\begin{abstract}
The goal of multi-label learning (MLL) is to associate a given instance with its relevant labels from a set of concepts. Previous works of MLL mainly focused on the setting where the concept set is assumed to be fixed, while many real-world applications require introducing new concepts into the set to meet new demands. One common need is to refine the original coarse concepts and split them into finer-grained ones, where the refinement process typically begins with limited labeled data for the finer-grained concepts. To address the need, we formalize the problem into a special weakly supervised MLL problem to not only learn the fine-grained concepts efficiently but also allow interactive queries to strategically collect more informative annotations to further improve the classifier. The key idea within our approach is to learn to assign pseudo-labels to the unlabeled entries, and in turn leverage the pseudo-labels to train the underlying classifier and to inform a better query strategy. Experimental results demonstrate that our pseudo-label approach is able to accurately recover the missing ground truth, boosting the prediction performance significantly over the baseline methods and facilitating a competitive active learning strategy.
\end{abstract}

\section{Introduction}
Multi-label learning (MLL) is an important learning problem with a wide range of applications \citep{elisseeff2001,boutell2004learning,zhang2006}. 
Traditionally, the problem has often been tackled under the fully-supervised setting \citep{Tsoumakas2007MultiLabelCA,Zhang2014ARO}, where an annotated training set that consists of fully-labeled examples is required to learn an accurate multi-label classifier.
While the traditional setting covers the scenario where the label classes are fixed before learning, many real-world applications face different situations. One scenario that is common in many applications is the growing number of classes \citep{zhu2018}, where the growth splits high-level concepts to finer-grained ones \citep{deng2014}.
For example, the set of classes might start from high-level concepts such as $\{$Animal, $\ldots$, Food $\}$, and then grow to include finer-grained concepts like $\{$Cat, $\ldots$, Dog, $\ldots$, Apple, $\ldots$, Banana$\}$. 
In fact, this hierarchical label collection process is considered by \citet{deng2014} as a scalable and efficient way to build large benchmark multi-label data sets, such as the MS COCO data set \citep{lin2014}.
Typical applications may have collected sufficient number of labeled data for learning the high-level concepts in a fully supervised manner, but it can be challenging for the applications to efficiently adapt the classifier from the high-level (coarse-grained) concepts to the finer-grained ones.
Conquering such \textit{refinement} challenge calls for two components: one is an effective learning model to exploit the fine-grained labels that have been collected, and the other is a strategic algorithm to actively collect a few more fine-grained and informative labels within a limited budget. In this work, we approach the problem, which we term \textit{Active Refinement Learning} for multi-label learning, by first focusing on the design of the former component---learning an accurate fine-grained classifier with only limited supervision. Having an underlying learner that could most effectively exploit the limited labels, we then design an active learning method tailored to couple the proposed learner. Finally, combining the two components, we present a total algorithm solution to most efficiently tackle the refinement problem.

\begin{figure}[!t]
\centering
\begin{minipage}{0.4\textwidth}
  \centering
  \includegraphics[width=1\linewidth]{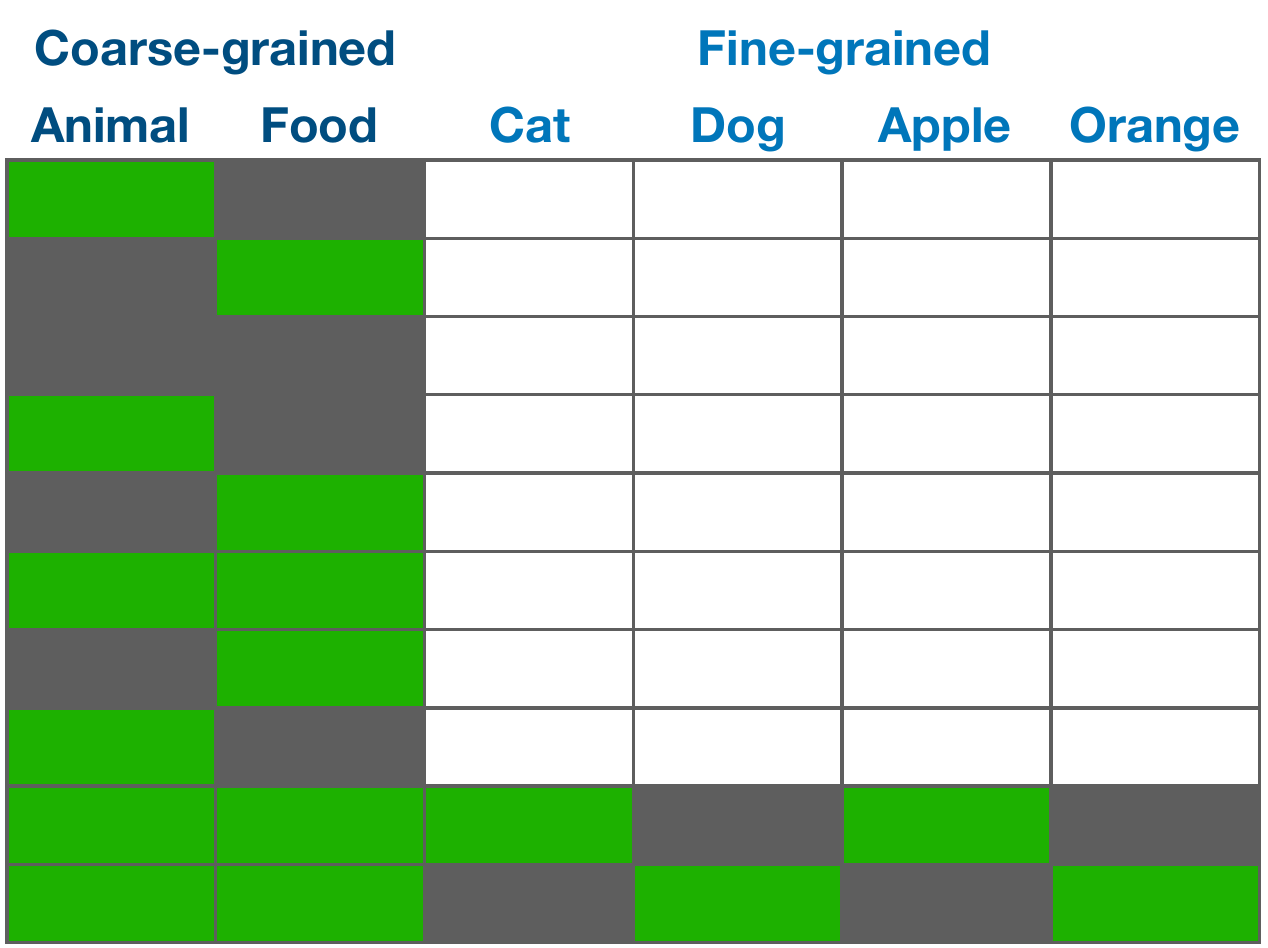}
\end{minipage}
\qquad
\begin{minipage}{0.4\textwidth}
  \centering
  \includegraphics[width=1\linewidth]{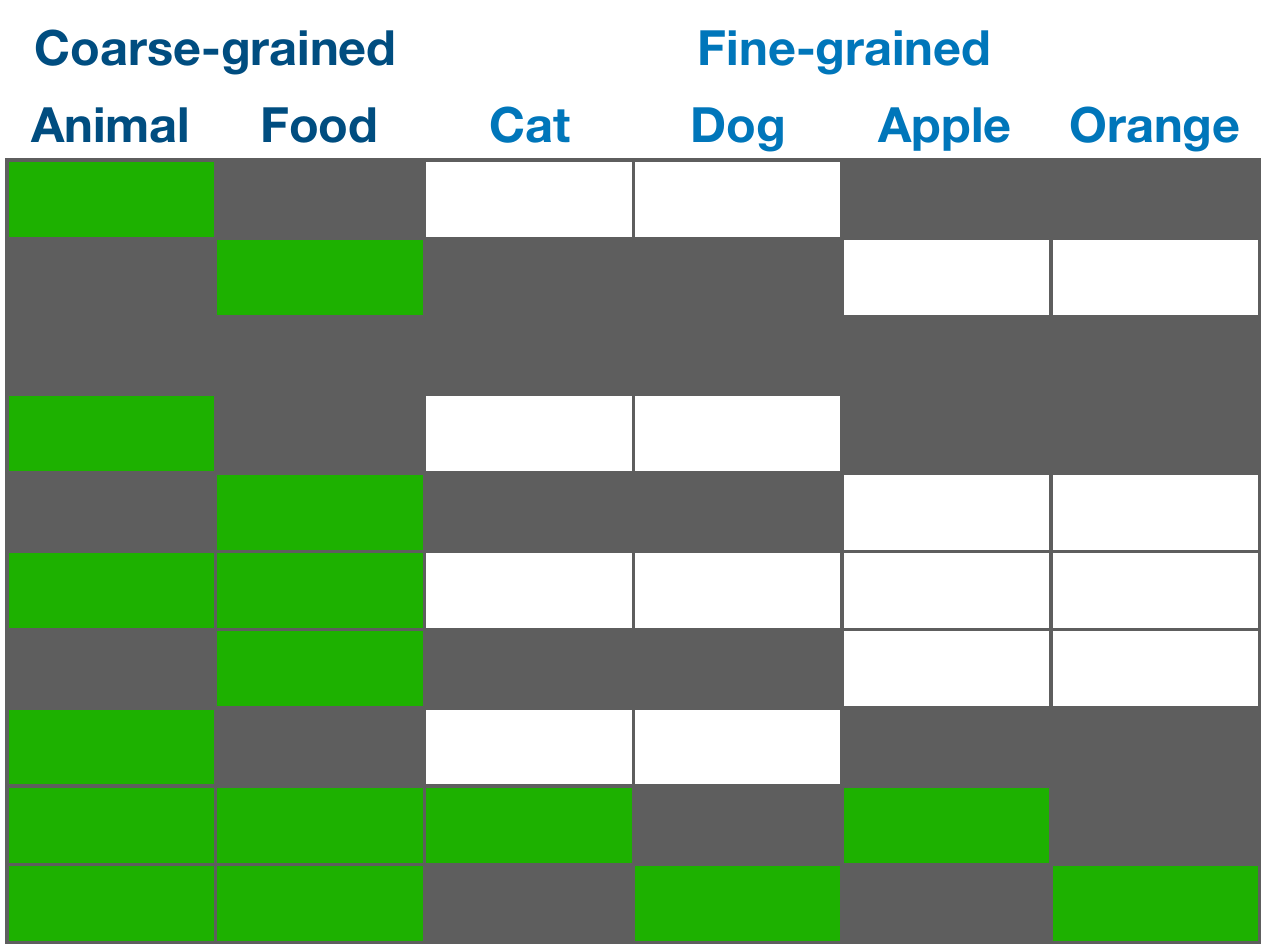}
\end{minipage}
\caption{Left: Original annotation received. Right: Annotation deduced from label hierarchy. Each row denotes the label vector of an example, where color green, gray and white correspond to relevant, irrelevant and unknown labels.}
\label{fig:supervision}
\end{figure}

\paragraph{Refinement Learning with Limited Supervision} 
To learn the newly introduced fine-grained concepts efficiently, an indispensable key is to have a learner that could maximally exploit the limited information on hand. In this work, we assume that the refinement process starts with a very limited number of fine-grained labels to warm start the classifier. In particular, the refinement starts with the model receiving a data set that contains all the coarse-grained labels and a few fine-grained ones, as shown in the left-hand side of Figure~\ref{fig:supervision}. Then, the problem of constructing a predictive fine-grained model with the presented data set falls under the big umbrella of weakly supervised learning \citep{zhou2018}. Specifically, when we focus on leveraging the coarse-grained labels to build a fine-grained classifier, the problem resembles learning with \textit{inexact supervision} considered by \citet{zhou2018}, where the coarse-grained labels are not in the exact form for the desired output and could only provide weak information about the target fine-grained labels. On the other hand, if we focus on using the fine-grained part of the labels to train the classifier, the problem can be viewed as a multi-label variant of learning with \textit{incomplete supervision} as some instances receive their exact fine-grained ground-truth labels whereas some do not have labels at all \citep{zhou2018}. While both the aforementioned problems have attracted much research attention, the combination of them (inexact and incomplete supervision) which our problem of interest can be cast as, has not yet been carefully investigated to the best of our knowledge.

There are some possible solutions in the literature that might be directly used to tackle the problem. Somehow those solutions are, however, not particularly designed for the problem and hence all carry some caveats. One possible solution is to take the hierarchical relationship between the coarse- and fine-grained labels to assign the ``knowingly irrelevant'' fine-grained labels, partially filling up the fine-grained label vectors as shown in the right-hand side of Figure~\ref{fig:supervision}. Then, those assigned labels can be fed in to a one-class multi-label learning model~\citep{Yu2017SelectionON,yu2017} to learn a fine-grained (irrelevance) classifier. The caveat of the solution is that the weak supervision from the relevant coarse-grained labels is not exploited. In fact, the distribution of the relevant fine-grained labels is typically not modeled in the original one-class setting~\citep{Yu2017SelectionON,yu2017}, but only through some heuristic assumption. Another solution is to view the unassigned labels as ``missing entries'' of the label vectors, and apply techniques to recover the missing entries via low-rank assumptions on the label distribution~\citep{goldberg2010,xu2013,yu2014}. Nonetheless, different from typical assumption made by this line of methods, the unassigned fine-grained labels in our problem are missing \textit{structurally} rather than randomly, making it hard to recover the missing entries with general low-rank assumptions.

In this work, we propose a new model that undertakes the challenges posed by inexact and incomplete supervision through a novel learning to learn method which jointly exploits the hierarchical relationship between the coarse- and fine-grained labels, as well as the benefits of all available data in hand.
The key idea within our model is to take into account all available information to learn the labeling assignments for the unlabeled entries, called \textit{pseudo-labels}, and use them to guide the decent direction of the parameter updates on the underlying classifier.
To seek for the best assignment of the pseudo-labels, conceptually, we hope the model trained according to the pseudo-labels could achieve the best classification performance on a set of disjoint validation data. Inspired by recent works in few-shot meta-learning literature \citep{sachin2017,ren2018meta,ren2018}, we treat the handful of data points that receive their fine-grained annotations as the validation set, and view the classification performance on the set as the meta-objective to optimize for. Finally, we adopt an iterative strategy that tunes the pseudo-labels locally on the fly at each gradient descent step of the model parameters to mitigate the computationally prohibiting expense of the meta-optimization problem \citep{Vicente1994BilevelAM,finn2017}.
Finally, we experimentally demonstrate that the proposed method not only assigns accurate pseudo-labels to the unknown entries but also enjoys significantly better performance than other methods for learning fine-grained classifiers under the limited supervision setting.

\paragraph{Query Strategy for Refinement Learning}
With a label-efficient classifier in hand, a natural follow-up  question is how to further improve the classifier's performance given a bit more annotation budget available. Answering this question is not new to the literature of Active Learning \citep{settles2009active}, where the goal is to strategically query the most informative labels such that the classifier trained on these examples could achieve the best performance. However, as the query strategies are often designed to complement the underlying learner, it is unclear how well off-the-shelf sampling strategies could fit with the proposed pseudo-label learning algorithm specially designed for the refinement setting. As a result, to take into account how the underlying learner are trained with pseudo-labels, we customized a new query strategy that follows the spirit of the classic uncertainty sampling strategy \citep{lewis1994a}, but designed to leverage more of the information provided by pseudo-labels. Specifically, our proposed strategy takes into account how the classifier would change after updated by the pseudo-labels. Unlike the classic vanilla uncertainty sampling method that queries the data lying closest to the current classifier's decision boundary or those having the highest prediction entropy \citep{lewis1994a,dagan1995committee,tong2001support,culotta2005reducing}, our proposed method quantifies the notion of uncertainty by the expected difference in the classifier's prediction before and after an update led by the estimated pseudo-labels. With the explicit consideration of pseudo-labels, we experimentally show that the proposed sampling strategy indeed better benefits the underlying learner comparing to classic query methods.

\paragraph{Contributions}
To summarize our contributions:
\begin{enumerate}
    \item We formalize the problem of \textit{active refinement} for multi-label learning to address a common setting faced by many real-world applications.
    \item We propose a pseudo-label method for efficiently learning a fine-grained classifier with limited supervision.
    \item We combine the pseudo-label learning method with a tailored query strategy to form a total solution for the problem considered.
\end{enumerate}

\begin{figure}[!t]
% \begin{wrapfigure}{r}{0.45\textwidth}
  \begin{center}
  \adjustbox{max width = .5\linewidth}{
    \begin{tikzpicture}[->,>=stealth',
level 1/.style={sibling distance=9em},
level 2/.style={sibling distance=3em},
level distance = 3em] 
\node [arn_n] {$r$}
    child{ node [arn_n, label=left:Animal] {$y_1$} 
            child{ node [arn_n, label=below:Cat] {$y_{11}$}
            }
            child{ node [arn_n, label=below:Dog] {$y_{12}$}
            }
            child {node {$\textbf{\dots}$}}
    }
    child{ node [arn_n, label=left:Food] {$y_2$}
            child{ node [arn_n, label=below:Apple] {$y_{21}$} 
            }
            child{ node [arn_n, label=below:Orange] {$y_{22}$}
            }
            child {node{$\textbf{\dots}$}}
		}
	child {node {$\textbf{\dots}$}}
; 
\end{tikzpicture}
}
  \end{center}
  \vspace{-5mm}
  \caption{Label Refinement from Coarse-to-Fine.}
  \label{fig:tree}
% \end{wrapfigure}
\end{figure}
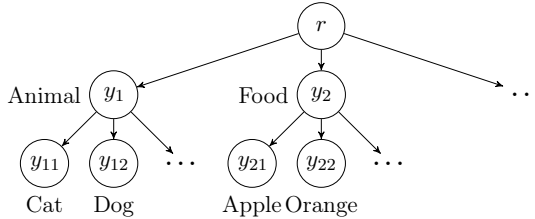

\section{Preliminaries}
\label{sec:preliminaries}
\subsection{Problem Setup: Refinement Learning for Multi-Label Learning}
Formally, we denote an instance by a feature vector $\mathbf{x} \in \mathbb{R}^d$, and its relevant labels by a bit vector $\mathbf{y} \in \{0, 1\}^K$ to indicate whether the labels in a pre-defined set $\mathcal{Y} = \{y_1, ..., y_K\}$ are relevant, i.e., $\mathbf{y}[k] = 1$ if and only if $y_k$ is relevant. In this work, rather than assuming that the set $\mathcal{Y}$ is fixed, we consider the problem of splitting the original high-level concepts into finer-grained ones, refining the label set of interest from $\mathcal{Y}_{\mathrm{c}} = \{y_1, ..., y_C\}$ into $\mathcal{Y}_{\mathrm{f}} = \{y_{11}, y_{12}, ..., y_{C1}, y_{C2}, ...\}$ as shown in Figure~\ref{fig:tree}.
Let $\mathbf{y}^\mathrm{c}$ and $\mathbf{y}^\mathrm{f}$ be the corresponding label vectors for $\mathcal{Y}_{\mathrm{c}}$ and $\mathcal{Y}_{\mathrm{f}}$ respectively. Assume that we receive a data set $\mathcal{D}_{\mathrm{tr}} = \{(\mathbf{x}_n, \mathbf{y}^{\mathrm{c}}_n)\}^{N}_{n=1}$ consisting of $N$ examples that are annotated only with the high-level (coarse-grained) labels, and an additional small warm-up set $\mathcal{D}'_{\mathrm{tr}} = \{(\mathbf{x}'_m, \mathbf{y}'^{\mathrm{f}}_m)\}^{M}_{m=1}$ of $M$ examples with their fine-grained labels annotated, our first goal is to leverage these examples to learn an accurate \textit{fine-grained} classifier $\Phi(\theta, \mathbf{x}): \mathbb{R}^d \to \{0, 1\}^K$ where $\theta$ is the model parameter and $K$ is the total number of fine-grained classes. 
% For simplicity, we denote the fine-grained label vectors in $\mathcal{D}_{tr}$ by a matrix form $\mathbf{Y}^f = [\mathbf{y}^f_1, ..., \mathbf{y}^f_N]^{\top}$.

% \paragraph{Tackling inexact supervision via label hierarchy}

\subsection{Related Work and Possible Existing Solutions}
\paragraph{Fully-Supervised Learning} A straightforward way to learn a fine-grained classifier is to utilize only the fully-annotated training examples in $\mathcal{D}'_{\mathrm{tr}}$ through standard supervised approaches. Nevertheless,
the small number of examples in this set might be unable to train a strong classifier. 
Moreover, it completely ignores the (weak) supervision provided by the abundant coarse labels.

\paragraph{Multi-Label Learning with Missing Labels}

One way to make use of the higher-level supervision on learning the fine-grained concepts is 
% to exploit the hierarchical relationship between the coarse- and fine-grained labels.
% A particular hierarchical relationship is
to leverage the hierarchical relationship 
where the irrelevance of a parent (coarse) concept implies the irrelevance of all of its children (fine) concepts. Leveraging the relationship, we are able to infer the corresponding entries in the fine-grained label vectors of the examples in $\mathcal{D}_{\mathrm{tr}}$,
making $\mathbf{Y}^{\mathrm{f}} = [\mathbf{y}^{\mathrm{f}}_1, ..., \mathbf{y}^{\mathrm{f}}_N]^\top$ partially observable, 
and reduce the original problem into a multi-label version of the negative-unlabeled learning problem~\citep{Niu2016TheoreticalCO,sakai2017} with very few positive examples, as shown in the right-hand side of Figure~\ref{fig:supervision}.

To tackle the reduced problem, one way is to treat the unknown fine-grained labels as missing entries, and apply MLL algorithms that can learn with the presence of missing labels \citep{goldberg2010,xu2013,yu2014}. \citet{yu2014} proposed a classic empirical risk minimization styled method LEML, attempting to solve the optimization problem that arrives at the model parameters which can most accurately recover the observed training labels. Roughly, their objective is formulated as:
\begin{align}
    \theta^{*} = \mathop{\arg\min}_{\theta}( \sum_{(i,j) \in \Omega}L(\Phi(\theta, \mathbf{x}_i)[j], [\mathbf{Y}^{\mathrm{f}}]_{i,j}) + \sum_{m=1}^M\sum_{k=1}^K L(\Phi(\theta, \mathbf{x}'_m)[k], \mathbf{y}'^{\mathrm{f}}_m)[k])),
\label{eq:leml}
\end{align}
where $\Omega$ is the set of indices of the observed entries in $\mathbf{Y}^{\mathrm{f}}$ and $L$ is a loss function that measures the discrepancy between the predicted and ground-truth labels. We use binary cross entropy loss for $L$ throughout the paper if not otherwise mentioned.
% works under the assumption that the observed entries could provide adequate relevant and irrelevant learning signals, our problem largely deviates from the assumption where the majority of the observed entries are irrelevant labels.
% Although the method works under the standard setting where the unknown entries are missing \texit{randomly}
% attempts to mitigate the missing label problem
% by applying low-rank assumption on the label distribution, it remains unclear how it would perform when applied in our problem where the unknown entries are missing \textit{structurely}.
From the objective, however, we note that only the observed entries contribute to the learning of model parameters, and the unobserved ones are basically ignored in the model training process. Unfortunately, in our setting, as the observed fine-grained labels are mostly deduced from the irrelevance of their parent labels, LEML is thus unable to exploit the weak supervision provided by the relevant coarse labels.

\paragraph{One-Class Multi-Label Learning}
% As most of our observed entries are of the same (negative) class
Another plausible direction to approach the reduced problem is through one-class multi-label learning methods \citep{Yu2017SelectionON,yu2017}. These methods are designed to handle the setup where the observed entries all hold the same values.
A common approach took in these methods is to assume the values of the unobserved entries to be the opposite class of the observed ones, and train a cost-sensitive classifier with different weights given to the observed and unobserved entries. Namely, assume that the observed entries are all negative (as in the deduced fine-grained matrix $\mathbf{Y}^{\mathrm{f}}$), the one-class methods optimize the following objective function:
\begin{align}
    \theta^{*} = \mathop{\arg\min}_{\theta} \sum_{(i,j) \in \Omega \cup \overline{\Omega} }w_{ij} \cdot L( \Phi(\theta, \mathbf{x}_i)[j], \mathbf{Y}_{ij}), \quad \text{where} \quad \mathbf{Y}_{ij} =
                \begin{cases}
                  \mathbf{Y}^f_{i, j} & (i, j) \in \Omega\\
                  1 & (i, j) \not\in \Omega\\
                \end{cases}
                ,
\label{eq:occ_obj}
\end{align}
where $w_{ij}$ is the corresponding weight on the loss of each entry, and is typically set to make $w_{ij} > w_{kl}$ for $(i,j) \in \Omega$ and $(k, l) \notin \Omega$, i.e., the observed entries have higher weights than the unobserved ones. Note that the unobserved entries are set to some default value (in this case, $1$) in the objective function.
Nonetheless, as the underlying ground truths for the missing entries are not necessarily the presumed class, without careful algorithm redesign or label distribution estimation, these methods may suffer from the introduced label bias that results in suboptimal performances \citep{Plessis2014AnalysisOL,Plessis2015ConvexFF,Kiryo2017PositiveUnlabeledLW}.

% \paragraph{Hierarchical Multi-Label Learning}

% \section{Learning from Limited Supervision}
\section{Active Refinement Learning for Multi-Label Learning} \label{sec:3}
\subsection{A Pseudo-Label Method for Refinement Learning}
% From Eq.~\ref{eq:leml}, we observe that different ways of treating the missing entries could exploit different degree of benefits, but at the same time carry different caveats.
% From the potential solutions, 
% we note that by treating the missing entries differently, we could exploit different degrees of benefits from them. 
% we observe that existing methods carry the caveats that the undetermined
% However, as existing methods all carry different caveats mentioned earlier, 
While existing solutions have developed different ways of treating the unknown entries during the learning process, they somehow do not delicately exploit the benefits of the unlabeled entries as mentioned in the previous section.
In light of this, we seek for a method that could more properly leverage the missing entries with a key assumption that: 
\textit{When the missing entries are all correctly recovered and used in the training process, the classifier learned could achieve the best performance}. Based on the assumption,
% To avoid either overlooking or simply presuming a default value for the missing entries, we are motivated to more correctly recover the unobserved entries with the aim that the recovered entries can guide to learn a better classifier.
% Towards this goal, we propose to follow the meta-learning paradigm where we treat the missing entries in $[\mathbf{y}^f_1, ..., \mathbf{y}^f_N]^\top$ as learnable parameters in our model, and utilize the classification performance on the small set $\mathcal{D}'_{tr}$ as the mete-objective to optimize for.
we attempt to find the best labeling assignment to the unknown entries, called pseudo-labels, which when the model is trained accordingly, can lead to best classification performance on the fine-grained concepts.
Towards this goal, we propose to use the few examples that receive their fully-annotated fine-grained labels in $\mathcal{D}'_{\mathrm{tr}}$ as a validation set to evaluate the classifier's performance on the fine-grained concepts. Formally, we aim to train our fine-grained classifier using the examples in $\mathcal{D}_{\mathrm{tr}}$ with a pseudo fine-grained label matrix $\mathbf{Y}^{\mathrm{pseudo}}$ where:

\begin{equation}
    [\mathbf{Y}^{\mathrm{pseudo}}]_{i, j} =
                \begin{cases}
                  [\mathbf{Y}^{\mathrm{f}}]_{i, j} & (i, j) \in \Omega\\
                  p_{ij} \in [0, 1] & (i, j) \not\in \Omega\\
                \end{cases},
\end{equation}
where every $p_{ij}$ is a pseudo label to be determined and the objective is:

\begin{equation}
\theta^*(\mathbf{Y}^{\mathrm{pseudo}}) = \mathop{\arg\min}_{\theta}\sum_{n=1}^N\sum_{k=1}^KL(\Phi(\theta, \mathbf{x}_n)[k], [\mathbf{Y}^{\mathrm{pseudo}}]_{n,k}).
\end{equation}
Note that with different pseudo-labels assigned to the missing entries, we arrive at different optimal model parameter $\theta^*(\mathbf{Y}^{\mathrm{pseudo}})$. From our assumption, the optimal assignment of the pseudo-labels should be based on the validation performance of the resulting classifier:

\begin{equation}
(\mathbf{Y}^{\mathrm{pseudo}})^* = \mathop{\arg\min}_{\mathbf{Y}^{\mathrm{pseudo}}}\sum_{m=1}^M\sum_{k=1}^KL(\Phi(\theta^*(\mathbf{Y}^{\mathrm{pseudo}}), \mathbf{x}'_m)[k], \mathbf{y}'^{\mathrm{f}}_m[k]).
\label{eq:y*}
\end{equation}
% minimizing the fine-grained classification loss on the examples in $\mathcal{D}'_{tr}$.
% Intuitively, we can view the pseudo-labels as the hyperparameters of the model, and we attempt to seek the best assignment of them such that the trained model can make the most accurate classification on the validation dataset, which in our case is $\mathcal{D}'_{tr}$ whose fine-grained labels are fully-annotated.
However, solving Eq.~\ref{eq:y*} to find the optimal pseudo-label assignment requires a computationally prohibiting bi-level optimization procedure \citep{Vicente1994BilevelAM}. To conquer the optimization challenge, inspired by recent works in meta-learning literature \citep{finn2017,ren2018}, we attempt to tackle the problem with an iterative approach which dynamically find the best pseudo-label assignments locally at each optimization step.
Specifically, consider a typical gradient descent update step:
\begin{equation}
\theta^{t+1} = \theta^{t} - \alpha \left.\nabla \sum_{n=1}^N\sum_{k=1}^K L(\Phi(\theta, \mathbf{x}_n)[k], [\mathbf{Y}^{\mathrm{pseudo}}]_{n,k}) \right|_{\theta=\theta^t},
\label{eq:theta_update}
\end{equation}
where $\alpha$ is the step size and $t$ is the current timestep.
Then, at each iteration $t$, we aim to learn the pseudo-label assignment which leads to the model parameters that minimize the validation loss after a single update step:

\begin{equation}
(\mathbf{Y}^{\mathrm{pseudo}})^*_t = \mathop{\arg\min}_{\mathbf{Y}^{\mathrm{pseudo}}}\sum_{m=1}^M\sum_{k=1}^KL(\Phi(\theta_{t+1}, \mathbf{x}'_m)[k], \mathbf{y}'^{\mathrm{f}}_m[k]).    
\label{eq:best_single}
\end{equation}
Solving Eq.~\ref{eq:best_single} at each timestep $t$ could, nevertheless, still be very expensive. As a result, we propose a simple approximation of $(\mathbf{Y}^{\mathrm{pseudo}})^*_t$ by looking at the gradient direction (sign) of the validation loss wrt. the pseudo-labels. Particularly, we assign pseudo-labels at timestep $t$ by:
\begin{align}
[\mathbf{Y}^{\mathrm{pseudo}}_t]_{i,j} =  \mathbb{I}(\frac{\partial}{\partial [\mathbf{Y}^{\mathrm{pseudo}}]_{i,j}}\sum_{m=1}^M\sum_{k=1}^KL(\Phi(\theta_{t+1}, \mathbf{x}'_m)[k], \mathbf{y}^{\mathrm{f}}_m[k]) \leq 0), \quad
\forall (i, j) \notin \Omega,
\label{eq:grad_dir}
\end{align}
where $\mathbb{I}$ is the indicator function.
After obtaining the newly assigned pseudo-labels $\mathbf{Y}^{\mathrm{pseudo}}_t$, which could be viewed as approximately the best label assignment locally at step $t$, we could utilize them to guide real parameter update on $\theta^t$. We outline the complete learning algorithm with pseudo-labels in Algorithm~\ref{algo:refinement_learning}.

\SetKwInput{KwInput}{Input}
\begin{algorithm}[!t]
\caption{Refinement Learning with Pseudo-Labels}
\label{algo:refinement_learning}
\KwInput{$\mathcal{D}_{\mathrm{tr}}$ and $\mathcal{D}'_{\mathrm{tr}}$}
Initialize $\theta^0$ and $\mathbf{Y}^{\mathrm{pseudo}}$;

 \For{$t = 0 \ldots T-1$}{
    Sample an example $\mathbf{x}_n$ and its currently assigned pseudo-labels $\mathbf{Y}^{\mathrm{pseudo}}[n, :]$;
    
    Perform pseudo-update step on $\theta^t$ and obtain a temporary $\theta^{t+1}$ by Eq.~\ref{eq:theta_update};
    
    Calculate partial derivatives of $\mathbf{Y}^{\mathrm{pseudo}}$ and obtain $\mathbf{Y}^{\mathrm{pseudo}}_{t}$ by Eq.~\ref{eq:grad_dir};
    
    Perform real parameter update $\theta^{t+1} \leftarrow \theta^t$ using the above obtained $\mathbf{Y}^{\mathrm{pseudo}}_{t}$;
  
 }
\end{algorithm}

\subsection{Active Refinement Learning with Pseudo-Label}
\label{active_pseudo}
In the previous section, we have discussed about how to design a strategic algorithm that could learn to predict accurately given only a handful of supervision. In practice, however, in addition to an effective algorithm that can sufficiently exploit all the information residing in the limited data, another critical component is how to efficiently gather new labels that could in turn boost the prediction accuracy the most if we are allowed a little bit more budget on obtaining fine-grained annotations. As a result, in this section, we complement the proposed algorithm by further leveraging the concept of pseudo-label to design an efficient active learning strategy that together yields a total solution for the proposed problem of active refinement learning for multi-label classification.

To start with, consider that we are initially given $M$ fully-annotated examples. Our goal of the active learning algorithm is to query as few as possible the ground truth of the unknown fine-grained labels in the remaining $N$ examples, i.e. the white entries in the right-hand side of Figure~\ref{fig:supervision}, such that the classifier could learn to achieve the highest accuracy. Of course, one naive query strategy is to simply query the unknown labels \textit{randomly} without taking into account the data distribution and the current model behavior. Another commonly adopted strategy is \textit{uncertainty sampling} which has an underlying assumption that the most informative data are those that the model is most uncertain with, and thus suggests to query the labels that lie closest to the decision boundary of the current model \citep{lewis1994a}. While comparing to the random baseline, uncertainty sampling benefits from considering the current model behavior and seek to improve the model's prediction confidence by selecting to query the labels with the highest prediction uncertainty. A main caveat of such strategy is that it might lead to over-confidence of the model, resulting in suboptimal prediction accuracy in the end. For example, uncertainty sampling is prone to ignore the examples that lie far away from the current decision boundary which in fact have the opposite ground-truth labels to what the model predicts them to be \citep{Huang2014ActiveLB}. In such cases, the model ultimately only optimizes for the accuracy in a rather local region around its initial decision boundary. To look further into such problem, the reason why uncertainty sampling could not take into account the existence of overly confident predictions (which are actually wrong) might be due to the fact that the model's prediction itself seems to be the only guess of the ground-truth label of the data. If there is other more precise way to estimate the ground truth of the labels, we might be able to take into account this information and in turn adjust our query strategy. This idea then naturally leads to leveraging our proposed pseudo-labels to design a more sophisticated active learning strategy.

Our proposed method closely follows the concept of uncertainty sampling in the sense that we select the most \textit{uncertain} labels to query. However, instead of determining the uncertainty merely based on its distance to the current decision boundary, or similarly, the entropy of the predictive distribution, we quantify the notion of uncertainty by measuring how large the label prediction would change after the model is updated accordingly. But how can we know how the model will change before actually querying and obtaining the ground truth labels? This thus leads to the use of the proposed pseudo-labels. Specifically, in each active learning iteration with current model parameter $\theta^t$, we first obtain each unknown entry's label prediction by:
\begin{align}
    \hat{y}_{i,j} = \Phi(\theta^t, \mathbf{x}_i)[j], \quad \forall (i, j) \notin \Omega.
    \label{eq:current_y}
\end{align}
Then, we apply Eq.~\ref{eq:grad_dir} to obtain the pseudo-labels of the unknown entries and perform a pseudo parameter update by using the pseudo-labels as the ground truths, and arrive at model parameter $\hat{\theta^{t}}$. By looking ahead on the updated parameters $\hat{\theta}^t$, we could obtain new predictions for the unknown entries by:
\begin{align}
 \hat{y}'_{i,j} = \Phi(\hat{\theta}^t, \mathbf{x}_i)[j], \quad \forall (i, j) \notin \Omega.
 \label{eq:update_y}
\end{align}
Finally, we compute the cross-entropy between $\hat{y}_{i,j}$ and $\hat{y}'_{i,j}$ to measure the uncertainty of each entry. Our algorithm then selects to query for the ground truth label of $[\mathbf{Y}^{\mathrm{f}}]_{i,j}$, where the entry $(i, j)$ has the highest uncertainty:
\begin{align}
     (i, j)  = \arg\max_{i, j} -(\hat{y}_{i,j}\log\hat{y}'_{i,j} + (1-\hat{y}_{i,j})\log(1-\hat{y}'_{i,j}))
\label{eq:query}
\end{align}
We note that our method also resembles the design of \textit{Expected Model Change} \citep{Settles2007MultipleInstanceAL,settles2009active}, in a sense that we utilize the pseudo-labels to look a step ahead into how the model will change (and hence its prediction) and leverage that information to inform our query selection.
We outline the complete algorithm for active refinement learning with pseudo-labels in Algorithm~\ref{algo:active_refinement_learning}.

\SetKwInput{KwInput}{Input}
\begin{algorithm}[!t]
\caption{Active Refinement Learning with Pseudo-Labels}
\label{algo:active_refinement_learning}
\KwInput{$\mathcal{D}_{\mathrm{tr}}$, $\mathcal{D}'_{\mathrm{tr}}$, and Query Budget $B$}

Initialize $\theta^0$;

\For{$b = 0 \ldots B-1$}{

Compute the current prediction of the unknown entries with $\theta^t$ by Eq.~\ref{eq:current_y};

Calculate pseudo-labels for the unknown entries according to Eq.~\ref{eq:grad_dir};

Perform pseudo-update $\hat{\theta}^t \leftarrow \theta^t$ with the above pseudo-labels;

Compute the look-ahead prediction of the unknown entries with $\hat{\theta}^t$ by Eq.~\ref{eq:update_y};

Query for the ground truth label of $[\mathbf{Y}^{\mathrm{f}}]_{i,j}$ according to Eq.~\ref{eq:query};

Initialize $\mathbf{Y}^{\mathrm{pseudo}}$;

Learn the classifier with Algorithm~\ref{algo:refinement_learning};
 }
\end{algorithm}

\section{Experiments}
To justify the effectiveness of the proposed methods, we test our methods on two multi-label image datasets, MS COCO \citep{lin2014} and MirFlickr \citep{huiskes08the}.
The MS COCO data set contains in total 82081 training instances and 40137 testing instances, with 12 coarse-grained concepts and 80 fine-grained concepts; the MirFlickr data set consists of 14753 training instances and 9828 testing instances, with 10 and 38 coarse- and fine-grained concepts respectively.
In the experiments, we utilize the image features of size 2048 extracted with a deep residual network, ResNet-50 \citep{he2016}, pre-trained on the ImageNet data set \citep{deng2009}.

\begin{table}[!t]
\small
\centering
\caption{Precision@$k$ for different methods at different ratios of M/(M+N) on MS COCO.}
\begin{tabular}{cccccc}
\toprule
Ratio & Metric &  \multicolumn{4}{c}{Methods} \\
\cmidrule(lr){1-1} \cmidrule(lr){2-2} \cmidrule(lr){3-6}
$\log_2 (\frac{M}{M+N})$ & P@$k$ & Fully-Supervised  & LEML & One-Class Classification & Ours\\

\midrule

\multirow{3}{*}{-10} & P@$1$ & 0.5959 & 0.4689 & 0.5515 & \textbf{0.6633}\\
 & P@$3$ & 0.3406 & 0.2499 & 0.2593 & \textbf{0.4024}\\
 & P@$5$ & 0.2487 & 0.1785 & 0.2203 & \textbf{0.3083}\\
 
 \midrule
 
\multirow{3}{*}{-9} & P@$1$ & 0.6527 & 0.6566 & 0.6105 & \textbf{0.7217}\\
 & P@$3$ & 0.3822 & 0.3890 & 0.3477 & \textbf{0.4391}\\
 & P@$5$ & 0.2802 & 0.2883 & 0.2615 & \textbf{0.3227}\\
 
 \midrule
 
 \multirow{3}{*}{-8} & P@$1$ & 0.6797 & 0.6128 & 0.7051 & \textbf{0.7410}\\
 & P@$3$ & 0.4223 & 0.3214 & 0.4021 & \textbf{0.4612}\\
 & P@$5$ & 0.3127 & 0.2252 & 0.2867 & \textbf{0.3482}\\
 
 \midrule
 
 \multirow{3}{*}{-7} & P@$1$ & 0.7088 & 0.6654 & 0.7408 & \textbf{0.7705}\\
 & P@$3$ & 0.4369 & 0.3858 & 0.4379 & \textbf{0.4872}\\
 & P@$5$ & 0.3220 & 0.2810 & 0.3106 & \textbf{0.3607}\\
 
 \midrule
 
 \multirow{3}{*}{-6} & P@$1$ & 0.7631 & 0.7543 & 0.7564 & \textbf{0.7792}\\
 & P@$3$ & 0.4653 & 0.4448 & 0.4165 & \textbf{0.4922}\\
 & P@$5$ & 0.3387 & 0.3211 & 0.2946 & \textbf{0.3637}\\

\bottomrule

\end{tabular}

\label{exp:table1}

\end{table}

\begin{table}[!t]
\small
\centering
\caption{Precision@$k$ for different methods at different ratios of M/(M+N) on MirFlickr.}
\vspace{2mm}
\begin{tabular}{cccccc}
\toprule
Ratio & Metric &  \multicolumn{4}{c}{Methods} \\
\cmidrule(lr){1-1} \cmidrule(lr){2-2} \cmidrule(lr){3-6}
$\log_2 (\frac{M}{M+N})$ & P@$k$ & Fully-Supervised  & LEML & One-Class Classification & Ours\\

\midrule

\multirow{3}{*}{-10} & P@$1$ & 0.6033 & 0.4964 & 0.5724 & \textbf{0.8595}\\
 & P@$3$ & 0.5045 & 0.3675 & 0.4509 & \textbf{0.6617}\\
 & P@$5$ & 0.4255 & 0.3058 & 0.4154 & \textbf{0.5532}\\
 
 \midrule
 
\multirow{3}{*}{-9} & P@$1$ & 0.6349 & 0.5699 & 0.5760 & \textbf{0.8849}\\
 & P@$3$ & 0.5180 & 0.4768 & 0.4655 & \textbf{0.6845}\\
 & P@$5$ & 0.4476 & 0.4214 & 0.4170 & \textbf{0.5675}\\
 
 \midrule
 
 \multirow{3}{*}{-8} & P@$1$ & 0.7184 & 0.6641 & 0.6385 & \textbf{0.9025}\\
 & P@$3$ & 0.5829 & 0.5273 & 0.5035 & \textbf{0.6951}\\
 & P@$5$ & 0.4975 & 0.4529 & 0.4400 & \textbf{0.5747}\\
 
 \midrule
 
 \multirow{3}{*}{-7} & P@$1$ & 0.7676 & 0.6875 & 0.6533 & \textbf{0.9107}\\
 & P@$3$ & 0.6200 & 0.5593 & 0.5091 & \textbf{0.7126}\\
 & P@$5$ & 0.5254 & 0.4797 & 0.4373 & \textbf{0.5801}\\
 
 \midrule
 
 \multirow{3}{*}{-6} & P@$1$ & 0.7990 & 0.7533 & 0.6704 & \textbf{0.9094}\\
 & P@$3$ & 0.6523 & 0.6064 & 0.5248 & \textbf{0.7244}\\
 & P@$5$ & 0.5469 & 0.5125 & 0.4437 & \textbf{0.5932}\\

\bottomrule

\end{tabular}

\label{exp:table2}

\end{table}

\subsection{Refinement Learning with Limited Fine-Grained Labels}
In the first part of the experiments, we validate how the proposed pseudo-label approach could be used to learn an accurate classifier with only limited fine-grained supervision. We compare the proposed method with three baselines, namely, (a) a standard fully-supervised (FS) learning model, (b) LEML, a classic approach in handling typical missing-label setup \citep{yu2014}, and (c) an exemplar method that tackles the problem of one-class classification (OCC) \citep{yu2017}. 
As discussed in Section~\ref{sec:preliminaries}, these selected baselines serve as representative methods in different families that can be more easily adapted to approach the refinement problem considered. We thus choose to restrict our focus on benchmarking the proposed solution with these representative methods, and leave further exploration and comparison with their more complicated variants as future direction.
For fair comparison, we fix the base model architecture to be a multi-layer fully-connected neural network across all methods considered such that we could focus on the effect of different training strategies. Specifically, for the base model, we employ a network consists of an input layer with size 2048, two hidden layers each with size 512, and an output layer of size corresponding to the total number of fine-grained labels of interest. 

To evaluate the performances of the learned multi-label classifiers, we utilize the commonly adopted criterion \textit{Precision@$k$} (P@$k$) \citep{Yen2017PPDsparseAP,wydmuch2018ANG,menon2019MultilabelRW}:
\begin{align}
    \text{Precision@}k(\hat{\mathbf{y}}, \mathbf{y}) = \frac{1}{k} \sum_{l \in \texttt{rank}_k(\hat{\mathbf{y}})} \mathbf{y}[l],
\end{align}
where $\hat{\mathbf{y}} \in \mathbb{R}^K$ is the predicted score vector, $\mathbf{y} \in \{0, 1\}^K$ is the ground truth label vector, and $\texttt{rank}_k(\hat{\mathbf{y}})$ returns the top-$k$ largest indices of $\hat{\mathbf{y}}$.
Precision@$k$ measures a multi-label model's ability to correctly rank the relevance of each label with respect to a given instance. Evaluation on the ranking avoids the requirement of selecting some hard threshold to bipartite relevant and irrelevant labels, where the best threshold value could vary drastically across different methods. We provide results of precision@$k$ at different values of $k$, which in essence correspond to the average precision.
We fine-tune the hyper-parameters for all methods using a held-out validation set, and report the average performance of each method over 10 different runs.

\begin{figure}[!t]
\centering
\begin{minipage}{.495\textwidth}
  \hfill
  \includegraphics[height=12em]{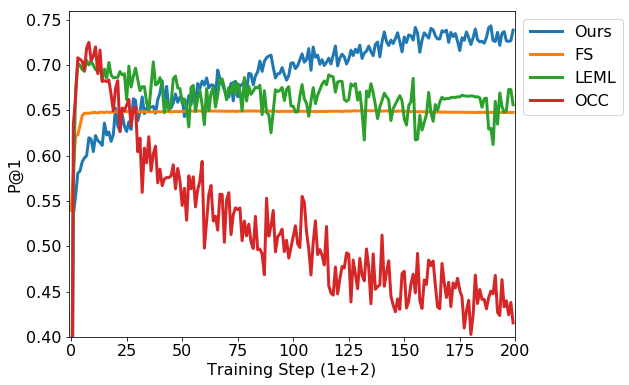}
  \caption{Learning curves of different methods on MS COCO. Ours corresponds to the proposed pseudo-label learning algorithm.}
  \label{fig:lc}
\end{minipage}
\begin{minipage}{.495\textwidth}
  \hfill
  \includegraphics[height=12em]{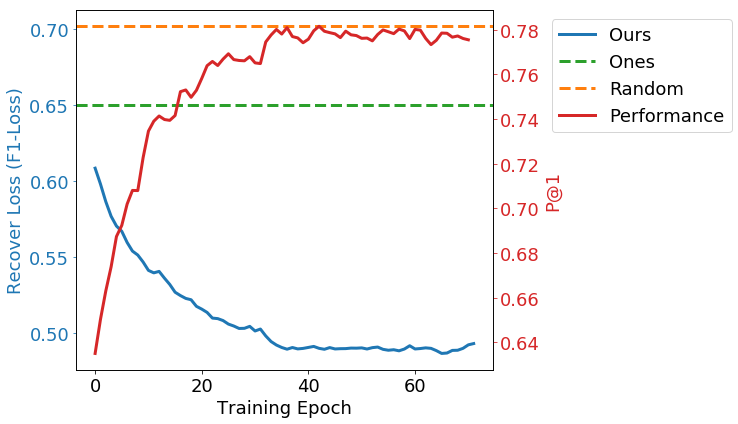}
%   \vspace{1mm}
  \caption{Recover rate of the unknown entries on MS COCO. Ours corresponds to the pseudo-label approach.}
  \label{fig:recover}
\end{minipage}
\end{figure}

\paragraph{Comparison with baseline methods}
In Table \ref{exp:table1} and \ref{exp:table2}, we show the results of different methods with varying size of fully-annotated warm-up set $\mathcal{D}'_{tr}$. Note that we especially focus on the regime when the ratio of $\mathcal{D}'_{tr} : \mathcal{D}_{tr}$ is particularly low to match the realistic scenario when only very scarce fine-grained supervision is available.
From the Tables, it can be seen that our method consistently achieves the best performances across different settings. It is worthwhile to note that although the standard fully-supervised approach does not leverage the partially labeled examples in $\mathcal{D}_{tr}$, it surprisingly outperforms the other two baseline methods in many cases. To investigate the reasons for so, we plot the learning curves of different methods in Figure~\ref{fig:lc}. From the figure, we see that although LEML and OCC achieve comparable, or even better, performances than the fully-supervised approach at the very beginning of the learning process, the two approaches then quickly suffers from overfitting that results in the performance drop. For LEML, we conjecture that the performance degrading comes from the overwhelming number of negative entries dominating the learning dynamic. And arguably, the severe overfitting of OCC results from the over-simple assumption on the missing entries which brings label bias into the learning objective.
% From the table, we also notice an interesting finding where the performance gap between our method and fully-supervisd approach becomes smaller as the size of fully-annotated examples grows larger. The result might suggest that our method is especially useful under very limited fine-grained supervision. 
%It might as well shed light on the future design ()

\paragraph{Recover rate of our method}
To understand the benefits of the pseudo-labels learned in our approach, we show how our method is capable of correctly recovering the missing entries, as well as the correlation between the recover rate and model performance. In Figure~\ref{fig:recover}, we plot the recover performance of the learned pseudo-labels measured by F1-loss (1 $-$ F1-score), and the horizontal bars are the corresponding F1-loss by simply treating all missing entries as ones and assigning them random labels. We can see from the figure that the pseudo-labels learned from our method could much more correctly recover the missing entries than the two naive baselines. In addition, there is a strong correlation between the recover rate and model classification performance. With more accurate assignment of pseudo-labels on the unknown entries, the trained model is able to achieve stronger classification performance.

\begin{figure}[!t]
\centering
  \includegraphics[width=0.55\linewidth]{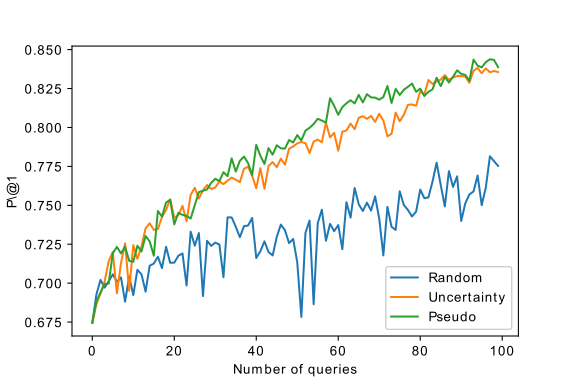}
  \caption{Progression of P@1 as more labels are queried and annotated by different active learning strategies using the pseudo-label learner as the base classifier on MS COCO. The green Pseudo curve corresponds to our proposed sampling strategy.}
  \label{fig:al_curve}
\end{figure}

\subsection{Active Refinement Learning}
In this part of the experiments, we evaluate the effectiveness of different sampling strategies when coupled with the proposed pseudo-label learning algorithm.
We compare our proposed strategy that leverages pseudo-labels in the query process as described in Section~\ref{active_pseudo} against other two baselines: (a) random sampling (RS), that is, to query for the ground truth uniformly over all unlabeled entries; and (b) uncertainty sampling (US), namely, to sample the unknown labels with largest model's classification entropy \citep{dagan1995committee}. In the experiments, we follow the common batch setting where we update the model periodically after querying a batch of labels \citep{shen2018deep,hu2018active}; in our setting, the learner queries 1000 labels per batch on MS COCO and queries 100 labels per batch on MirFlickr. Then, the classifier performs 1-epoch update on the data set augmented with the newly queried labels. To prevent the classifier overfits the initial data set, we reinitialize the model per 10 queries. 
% Also, to prevent the classifier underfits the data, we perform 10-epoch update on the classifier after initialization.
The performance of the classifier is evaluated with Precision@$k$ after each query.
As the goal is to achieve the best prediction accuracy with as least queries as possible, we plot the progression of the classifier performance as more labels are queried as shown in Figure~\ref{fig:al_curve}.
To compare between different progression curves, we follow common setup to aggregate the performance by computing the area under progression curves (AUC). We list the AUCs of different methods in Table~\ref{table:3}.

\paragraph{Comparison with baseline methods}
From Table \ref{table:3}, we observe that both uncertainty sampling and our proposed active learning strategy outperforms the naive random sampling baseline in most scenarios, which can mainly be attributed to the fact that random sampling ignores the relationship between the unlabeled data and the current model behavior. We also see a slightly better performance of our active learning strategy over the pure uncertainty sampling baseline. We conjecture such improvement comes from the utilization of pseudo-labels which allows our method to look ahead into how the model would change after a single iteration of update and further leverage this information to query the labels that would still remain highly uncertain to the model, instead of selecting the labels to query based directly on the current uncertainty level.

\begin{table}[!t]
\small
\centering
\caption{Area Under the Curve (AUC) of Precision@$k$ for different sampling strategies under different datasets.}
\label{table:3}
\adjustbox{max width = \textwidth}{
\begin{tabular}{ccccc}
\toprule
Base Classifier                &\multicolumn{4}{c}{Refinement Learning with Pseudo-label} \\ 
% &\multicolumn{3}{c}{Standard Fully-supervised}      \\
\midrule
Sampling strategies          &     & RS    & US    & Ours \\% & RS    & US    & ARL  \\
\midrule
                           & P@1 & 0.7246 & 0.7721 & \textbf{0.7782} \\%& 0.7780 & 0.7849 & \textbf{0.7866} 
%  \\
\cmidrule(l){2-5}
MS COCO                       & P@3 & 0.4494 & 0.4691 & \textbf{0.4727} \\%& \textbf{0.4726} & 0.4645 & 0.4725 
%  \\
\cmidrule(l){2-5}
                           & P@5 & 0.3400 & 0.3440 & \textbf{0.3482} \\%& \textbf{0.3472} & 0.3303 & 0.3415 
% \\
\midrule
                           & P@1 & 0.8822 & 0.8832 & \textbf{0.8843} 
 \\%& 0.8748 & 0.8362 & \textbf{0.8753} 
%  \\
\cmidrule(l){2-5}
MirFlickr                  & P@3 & 0.6766 & \textbf{0.6789} & 0.6785 
 \\%& \textbf{0.7136} & 0.6679 & 0.7103 
% \\
\cmidrule(l){2-5}
                           & P@5 & 0.5599 & \textbf{0.5624} & 0.5611 
 \\%& \textbf{0.5992} & 0.5521 & 0.5916 
%  \\
\bottomrule
\end{tabular}}
\end{table}

\section{Conclusion}
We formalize the problem of active refinement learning for multi-label classification to address a common need faced by many real-world applications.
We develop a novel pseudo-label learning method through a meta-learning strategy which could efficiently learn an accurate classifier with very limited fine-grained supervision. Moreover, we design a tailored active learning strategy for the pseudo-label learner to query for the most informative labels that further improve the classifier's performance.
Experimental results show that the proposed approaches compare favorably against existing possible methods, constituting a promising solution for the active refinement learning problem.

\bibliography{main}
\bibliographystyle{plainnat}

\end{document}